\pdfoutput=1

\documentclass{article}
\usepackage{times}
\usepackage[left=1.1in,right=1.1in,top=0.9in,bottom=0.9in]{geometry}
\date{}
\usepackage{amsmath}

\usepackage{hyperref}
\usepackage{url}
\usepackage{amsfonts}
\usepackage{amssymb}
\usepackage{graphicx}
\usepackage{color}
\usepackage{amsthm}

\newtheorem{proposition}{Proposition}
\newcommand{\x}{\boldsymbol{x}}\newcommand{\abs}[1]{\lvert#1\rvert}\newcommand{\bx}{\x}
\newcommand{\w}{\boldsymbol{w}}
\newcommand{\h}{\boldsymbol{h}}
\newcommand{\y}{\boldsymbol{y}}
\renewcommand{\v}{\boldsymbol{v}}
\renewcommand{\a}{\boldsymbol{a}}

\newcommand{\W}{\mathcal{W}_{\gamma}}

\newcommand{\expect}[2]{\ensuremath{\big\langle #1 \big\rangle_{#2}}}

\title{Wasserstein Training of Boltzmann Machines}

\author{
\\
\normalsize Gr\'egoire Montavon\\
\normalsize Department of Computer Science\\
\normalsize Technische Universit\"at Berlin\\
\texttt{\small gregoire.montavon@tu-berlin.de}
\\\\
\normalsize Klaus-Robert M\"uller\thanks{Also with the Department of Brain and Cognitive Engineering, Korea University.}\\
\normalsize Department of Computer Science\\
\normalsize Technische Universit\"at Berlin\\
\texttt{\small klaus-robert.mueller@tu-berlin.de}
\\\\
\normalsize Marco Cuturi\\
\normalsize Graduate School of Informatics\\
\normalsize Kyoto University\\
\texttt{\small mcuturi@i.kyoto-u.ac.jp}
}

\begin{document}

\maketitle

\begin{abstract}
The Boltzmann machine provides a useful framework to learn highly complex, multimodal and multiscale data distributions that occur in the real world. The default method to learn its parameters consists of minimizing the Kullback-Leibler (KL) divergence from training samples to the Boltzmann model. We propose in this work a novel approach for Boltzmann training which assumes that a meaningful metric between observations is given. This metric can be represented by the Wasserstein distance between distributions, for which we derive a gradient with respect to the model parameters. Minimization of this new Wasserstein objective leads to generative models that are better when considering the metric and that have a cluster-like structure. We demonstrate the practical potential of these models for data completion and denoising, for which the metric between observations plays a crucial role.

\end{abstract}

\section{Introduction}\label{sec-intro}

Boltzmann machines \cite{DBLP:journals/cogsci/AckleyHS85} are powerful generative models that can be used to approximate a large class of real-world data distributions, such as handwritten characters \cite{DBLP:journals/neco/Hinton02}, speech segments \cite{DBLP:conf/nips/DahlRMH10}, or multimodal data \cite{DBLP:journals/jmlr/SrivastavaS14}. Boltzmann machines share similarities with neural networks in their capability to extract features at multiple scales, and to build well-generalizing hierarchical data representations \cite{DBLP:journals/jmlr/SalakhutdinovH09,DBLP:series/lncs/MontavonM12b}. The restricted Boltzmann machine (called RBM) is a special type of Boltzmann machine defining a probability distribution over a set of $d$ binary observable variables whose state is represented by the vector $\x \in \{0,1\}^d$ and a set of $h$ explanatory variables, also binary. The distribution of the RBM can always be written in marginalized form as
$$
p_\theta(\x) = {\textstyle \frac{1}{Z_\theta}} e^{-F_\theta(\x)}
$$
where the function $F_\theta(\x)$ is called the free energy and is parameterized by a vector of parameters $\theta$. $Z_\theta$ is called the partition function and normalizes the distribution $p_\theta$ to $1$. Given an empirical probability distribution $\hat p(\x) = \frac1N \sum_{n=1}^N \delta_{\x_n}$ where $(\x_n)_n$ is a list of $N$ observations in $\{0,1\}^d$, an RBM can be trained using information-theoretic divergences (see for example \cite{DBLP:journals/jmlr/MarlinSCF10}) by minimizing with respect to $\theta$ a divergence $\Delta(\hat{p},p_\theta)$ between the sample empirical measure $\hat p$ and the modeled distribution $p_\theta$:
\begin{equation}\label{eq:mll}
	\min_{\theta\in\Theta} \Delta(\hat{p} , p_\theta).
\end{equation}
When $\Delta$ is for instance the KL divergence, this approach results in the well-known Maximum Likelihood Estimator (MLE), which yields gradients for the $\theta$ of the form
\begin{equation}
\nabla_\theta \mathrm{KL} (\hat p \,\|\, p_\theta) = \big\langle \nabla_\theta F_\theta(\x) \big\rangle_{\hat p} - \big\langle \nabla_\theta F_\theta(\x) \big\rangle_{p_\theta},
\label{eq:kl-gradient}
\end{equation}
where the bracket notation $\langle \cdot \rangle_p$ indicates an expectation with respect to $p$. The KL gradient involves the mean of the gradient of $F_\theta$ evaluated on observations, contrasted by its expectation under $p_\theta$. Alternative choices for $\Delta$ are the Bhattacharrya/Hellinger and Euclidean distances between distributions, or more generally $F$-divergences or $M$-estimators \cite{huber2011robust}. They all result in comparable gradient terms, that try to adjust $\theta$ so that the fitting terms $p_\theta(\x_n)$ grow as large as possible.

We explore in this work a different scenario: what if $\theta$ is chosen so that $p_\theta(\x)$ is large, on average, when $\x$ is \emph{close} to a data point $\x_n$ in some sense, but not necessarily when $\x$ coincides \emph{exactly} with $\x_n$? To adopt such a geometric criterion, we must first define what closeness between observations means. In almost all applications of Boltzmann machines, such a metric between observations is readily available: One can for example consider the Hamming distance between binary vectors, or any other metric motivated by practical considerations\footnote{When using the MLE principle, metric considerations play a key role to define densities $p_\theta$, e.g. the reliance of Gaussian densities on Euclidean distances. This is the kind of metric we take for granted in this work.}. This being done, the geometric criterion we have drawn can be materialized by considering for $\Delta$ the Wasserstein distance \cite{villani09} (a.k.a. the Kantorovich or the earth mover's distance \cite{rubner1997earth}) between measures. This choice was considered in theory by \cite{Bassetti20061298}, who proved its statistical consistency, but was never considered practically to the best of our knowledge. This paper describes a practical derivation for a \emph{minimum Kantorovich distance estimator} \cite{Bassetti20061298} for Boltzmann machines, which can scale up to tens of thousands of observations. As will be described in this paper, recent advances in the fast approximation of Wasserstein distances \cite{cuturi2013sinkhorn} and their derivatives \cite{DBLP:conf/icml/CuturiD14} play an important role in the practical implementation of these computations.

\begin{figure}
\centering
\includegraphics[width=0.80\textwidth]{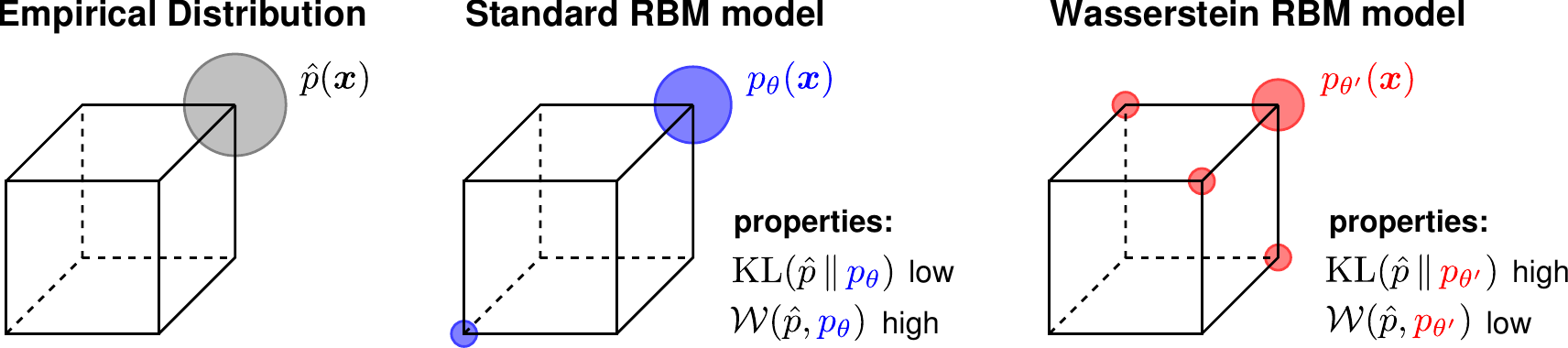}\\[-2mm]
\caption{Empirical distribution $\hat p(\x)$ (gray) defined on the set of states $\{0,1\}^d$ with $d=3$ shown next to two possible modeled distributions defined on the same set of states. The size of the circles indicates the probability mass allocated to each state. The first modeled distribution $p_\theta(\x)$ (blue) has low KL divergence and high Wasserstein distance from the empirical distribution. The second one $p_{\theta'}(\x)$ (red) has high KL divergence and low Wasserstein distance, and thus incorporates the desired metric.}
\label{figure:hypercube}
\end{figure}

Before describing this approach in detail, we would like to insist that measuring goodness-of-fit with the Wasserstein distance results in a considerably different perspective than that provided by a Kullback-Leibler/MLE approach. This difference is illustrated in Figure \ref{figure:hypercube}, where a probability $p_\theta$ can be close from a KL perspective to a given empirical measure $\hat p$, but far from the same measure $p$ in the Wasserstein sense. Conversely, a different probability $p_{\theta'}$ can miss the mark from a KL viewpoint but achieve a low Wasserstein distance to $\hat p$.

\section{A Practical Framework for Minimum Wasserstein Distance Estimation} \label{sec-wasserstein}

Consider two probabilities $p,q$ in $\mathcal{P}(\mathcal{X})$, the set of probabilities on $\mathcal{X}=\{0,1\}^d$. Namely, two maps $p,q:\mathcal{X}\rightarrow \mathbb{R}_+$ such that $\sum_{\x}p(\x)=\sum_{\x} q(\x)=1$, where we omit $\x\in\mathcal{X}$ under the summation sign. Consider a distance function $D:\mathcal{X}\times\mathcal{X}\rightarrow \mathbb{R}_+$ which satisfies for any triplet $\x,\x',\x''\in\mathcal{X}$ the triangle inequality $D(\x,\x'')\leq D(\x,\x') + D(\x',\x'')$ and $D(\x,\x)=0$. Given a constant $\gamma\geq 0$, the $\gamma$-smoothed Wasserstein distance \cite{cuturi2013sinkhorn} is equal to
\begin{equation}\label{eq:primal}\W(p,q) = \min_{\pi\in \Pi(p,q)} \langle D(\x,\x')\rangle_\pi -\gamma H(\pi),\end{equation}
where $\Pi(p,q)$ is the set of joint probabilities $\pi$ on $\mathcal{X}\times\mathcal{X}$ such that $\sum_{\x'} \pi(\x,\x')=p(\x)$, $\sum_{\x} \pi(\x,\x')=q(\x')$ and $H(\pi)=-\sum_{\x\x'}\pi(\x,\x')\log \pi(\x,\x')$ is the Shannon entropy of $\pi$. This optimization problem, a strictly convex program, has an equivalent dual formulation \cite{DBLP:conf/icml/CuturiD14} which involves instead two real-valued functions $\alpha,\beta$ on $\mathcal{X}$ and which plays an important role in this paper:
\begin{equation}\label{eq:dual}\W(p,q) = \max_{\alpha,\beta\in \mathbb{R}^{\mathcal{X}}} \langle \alpha(\x) \rangle_p + \langle \beta(\x') \rangle_q -\gamma \sum_{\x\x'} e^{\frac{1}{\gamma}(\alpha(\x) + \beta(\x') -D(\x,\x'))-1}.\end{equation}

\paragraph{Smooth Wasserstein Distances} The ``true'' Wasserstein distance corresponds to the case where $\gamma=0$, when Equation~\eqref{eq:primal} is stripped of the entropic term. The reader will easily verify that it matches the usual linear program used to describe Wasserstein/EMD distances \cite{rubner1997earth}. When $\gamma\rightarrow 0$ in Equation~\eqref{eq:dual}, one also recovers the Kantorovich dual formulation, because the rightmost regularizer converges to the indicator function of the feasible set of the dual optimal transport problem, $\alpha(\x)+\beta(\x')\leq D(\x,\x')$. We consider in this paper the case $\gamma>0$ because it was shown in \cite{cuturi2013sinkhorn} to considerably facilitate computations, and in \cite{DBLP:conf/icml/CuturiD14} to result in a divergence $\W(p,q)$ which, unlike the case $\gamma=0$, is \emph{differentiable} w.r.t to the first variable. Looking at the dual formulation in Equation~\eqref{eq:dual}, one can see that this gradient is equal to $\alpha^\star$, the centered optimal dual variable (the centering step for $\alpha^\star$ ensures the orthogonality with respect to the simplex constraint). 

Sensitivity analysis gives a clear interpretation to the quantity $\alpha^\star(\x)$: It measures the \emph{cost} for each unit of mass placed by $p$ at $\x$ when computing the Wasserstein distance $\W(p,q)$. To decrease $\W(p,q)$, it might thus be favorable to transfer mass in $p$ from points where $\alpha(\x)$ is high to place it on points where $\alpha(\x)$ is low. This idea can be used, by a simple application of the chain rule, to minimize, given a fixed target probability $p$, the quantity $\W(p_\theta,p)$ with respect to $\theta$.

\begin{proposition}
Let $p_\theta(\x) = \frac{1}{Z} e^{-F_\theta(\x)}$ be a parameterized family of probability distributions where $F_\theta(\x)$ is a differentiable function of $\theta\in\Theta$ and we write $G_\theta=\langle \nabla_\theta F_\theta(\x)\rangle_{p_\theta}$. Let $\alpha^\star$ be the centered optimal dual solution of $\W(p_\theta,p)$ as in Equation~\eqref{eq:dual}. The gradient of the smoothed Wasserstein distance with respect to $\theta$ is given by
\begin{equation}
\nabla_\theta \mathcal{W}_\gamma (p_\theta,p) =  \expect{\alpha^\star(\x)}{p_\theta} G_\theta - \expect{\alpha^\star(\x)\nabla_\theta F_\theta(\x))}{p_\theta}.
\label{eq:wasserstein-gradient}
\end{equation}
\begin{proof} This result is a direct application of the chain rule: We have
	$$
\nabla_\theta \mathcal{W}_\gamma (p_\theta,p) = \Big(\frac{\partial p_\theta}{\partial \theta}\Big)^{\!T} \, \frac{\partial \W(p_\theta,q)}{\partial p_\theta}.
	$$
As mentioned in \cite{DBLP:conf/icml/CuturiD14}, the rightmost term is the optimal dual variable (the Kantorovich potential) $\partial \W(p_\theta,q)/\partial p_\theta=\alpha^\star$. The Jacobian $\left(\partial p_\theta/\partial \theta \right)$ is a linear map $\Theta\rightarrow \mathcal{X}$. For a given $\x'$, $$\partial p_\theta(\x')/\partial \theta = p_\theta(\x')G_\theta -\nabla F_\theta(\x')p_\theta(\x').$$
As a consequence, $\big(\frac{\partial p_\theta}{\partial \theta}\big)^T \alpha^\star$ is the integral w.r.t. $\x'$ of 
the term above multiplied by $\alpha^\star(\x')$, which results in Equation~\eqref{eq:wasserstein-gradient}.
\end{proof}
\end{proposition}

\paragraph{Comparison with the KL Fitting Error} The target distribution $p$ plays a direct role in the formation of the gradient of  $\mathrm{KL} (\hat p \,\|\, p_\theta)$ w.r.t. $\theta$ through the term $\langle \nabla_\theta F_\theta(\x) \rangle_{p}$ in Equation~\eqref{eq:kl-gradient}. The Wasserstein gradient incorporates the knowledge of $p$ in a different way, by considering, on the support of $p_\theta$ only, points $\x$ that correspond to high potentials (costs) $\alpha(\x)$ when computing the distance of $p_\theta$ to $p$. A high potential at $\x$ means that the probability $p_\theta(\bx)$ should be lowered if one were to decrease $\W(p_\theta,p)$, by varying $\theta$ accordingly.

\paragraph{Sampling Approximation} The gradient in Equation \eqref{eq:wasserstein-gradient} is intractable, since it involves solving an optimal (smoothed) transport problem over probabilities defined on $2^d$ states. In practice, we replace expectations w.r.t $p_\theta$ by an empirical distribution formed by sampling from the model $p_\theta$ (e.g.\ the PCD sample \cite{DBLP:conf/icml/Tieleman08}). Given a sample $(\widetilde \x_{n})_n$ of size $\widetilde N$ generated by the model, we define $\hat p_\theta = \sum_{n=1}^{\widetilde N} \delta_{\widetilde \x_n}/\widetilde N$. The tilde is used to differentiate the sample generated by the model from the empirical observations. Because the dual potential $\alpha^\star$ is centered and $\hat p_\theta$ is a measure with uniform weights, $\langle \alpha^\star(\x) \rangle_{\hat p_\theta} = 0$ which simplifies the approximation of the gradient to
\begin{equation}
\widehat \nabla_{\theta} \mathcal{W}_\gamma(p_\theta,\hat p) = -{\textstyle \frac{1}{\widetilde N} \sum_{n=1}^{\widetilde N}} \hat\alpha^\star(\widetilde \x_n) \, \nabla_\theta F_\theta(\widetilde \x_n)
\label{eq:wasserstein-gradient-approximation}
\end{equation}
where $\hat\alpha^\star$ is the solution of the discrete smooth Wasserstein dual between the two empirical distributions $\hat{p}$ and $\hat p_\theta$, which have respectively supports of size $N$ and $\widetilde{N}$. In practical terms, $\hat\alpha^\star$ is a vector of size $\widetilde N$, one coefficient for each PCD sample, which can be computed by following the algorithm below \cite{DBLP:conf/icml/CuturiD14}. To keep notations simple, we describe it in terms of generic probabilities $p$ and $q$, having in mind these are in practice the training and simulated empirical measures $\hat{p}$ and $\hat p_\theta$.

\paragraph{Computing $\alpha^\star$} When $\gamma>0$, the optimal variable $\alpha^\star$ corresponding to $\W(p,q)$ can be recovered through the Sinkhorn algorithm with a cost which grows as the product $\abs{p}\abs{q}$ of the sizes of the support of $p$ and $q$, where $\abs{p}=\sum_{\x} 1_{p(\x)>0}$. The algorithm is well known but we adapt it here to our setting, see \cite[Alg.3]{DBLP:conf/icml/CuturiD14} for a more precise description. To ease notations, we consider an arbitrary ordering of $\mathcal{X}$, a set of cardinal $2^d$, and identify its elements with indices $1\leq i\leq 2^d$. Let $I=(i_1,\cdots,i_{\abs{p}})$ be the ordered family of indices in the set $\{i \,|\, p(i)>0\}$ and define $J$ accordingly for $q$. $I$ and $J$ have respective lengths $\abs{p}$ and $\abs{q}$.
Form the matrix $K=[e^{-D(i,j)/\gamma}]_{i\in I,j\in J}$ of size $\abs{p}$ and $\abs{q}$.
Choose now two positive vectors $u\in\mathbb{R}^{\abs{p}}_{++}$ and $v\in\mathbb{R}^{\abs{q}}_{++}$ at random, and repeat until $u,v$ converge in some metric the operations $u\leftarrow p/(Kv), v\leftarrow q/(K^Tu)$.
Upon convergence, the optimal variable $\alpha^\star$ is zero everywhere except for $\alpha^\star(i_a)=\log(u_a/\tilde{u})/\gamma$ where $1\leq a\leq \abs{p}$ and $\tilde{u}$ is the geometric mean of vector $u$ (which ensures that $\alpha^\star$ is centered).

\section{Wasserstein Training of a Restricted Boltzmann Machine} \label{sec-training}

The restricted Boltzmann machine (RBM) is a generative model of binary data that is composed of $d$ binary observed variables and $h$ binary explanatory variables. The vector $\x \in \{0,1\}^d$ represents the state of observed variables, and the vector $\y \in \{0,1\}^h$ represents the state of explanatory variables. The RBM associates to each configuration $\x$ of observed variables a probability $p_\theta(\x)$ defined as
$$
p_\theta(\x) = {\textstyle \frac{1}{Z_\theta} \sum_{\y \in \{0,1\}^h}} e^{-E_\theta(\x,\y)},
$$
where $E_\theta(\x,\y) = -\a^T \x - \sum_{j=1}^h y_j (\w_j^T \x + b_j)$ is called the energy and $\theta = (\a, \{\w_j,b_j\}_{j=1}^h)$ are the parameters of the RBM. These parameters must be learned from the data. Knowing the state $\x$ of the observed variables, the explanatory variables are independent Bernoulli-distributed with $\Pr(y_j=1 | \x) = \sigma(\w_j^T \x + b_j)$, where $\sigma$ is the logistic map $z \mapsto (1+e^{-z})^{-1}$. Conversely, knowing the state $\y$ of the explanatory variables, the observed variables on which the probability distribution is defined can also be sampled independently, leading to an efficient alternate Gibbs sampling procedure for $p_\theta$. In this RBM model, explanatory variables can be analytically marginalized, allowing us to rewrite the probability model as
$
p_\theta(\x) = {\textstyle \frac{1}{Z_\theta}} e^{-F_\theta(\x)},
$
where $F_\theta(\x) = -\a^T \x - \sum_j \log( 1+ \exp( \w_j^T \x + b_j))$ is the free energy associated to this model.

\paragraph{Wasserstein Gradient of the RBM} Having written the RBM in its free energy form, the Wasserstein gradient can be obtained by computing the gradient of $F_\theta(\x)$ and injecting it in Equation \eqref{eq:wasserstein-gradient-approximation}:
\begin{align*}
\widehat \nabla_{\w_j} \mathcal{W}_\gamma(p_\theta , \hat p)
= \big\langle \alpha^\star(\x) \, \sigma(z_j) \, \x \big\rangle_{\hat p_\theta},
\end{align*}
where $z_j = \w_j^T \x + b_j$. Gradients with respect to parameters $\a$ and $\{b_j\}_j$ can also be obtained by the same means. In comparison, the gradient of the KL divergence is given by 
$
\widehat \nabla_{\w_j} \mathrm{KL}(\hat p \,\|\, p_\theta)
= \big\langle \sigma(z_j) \, \x \big\rangle_{\hat p_\theta} - \big\langle \sigma(z_j) \, \x \big\rangle_{\hat p}.
$
While the Wasserstein gradient can in the same way as the KL gradient be expressed in a very simple form, the first one is not sum-decomposable. A simple manifestation of the non-decomposability occurs for $\widetilde N=1$ (smallest possible sample size): In that case, $\alpha(\widetilde \x_n)  = 0$ due to the centering constraint (see Section \ref{sec-wasserstein}), thus making the gradient zero.

\paragraph{Stability and KL Regularization} Unlike the KL gradient, the Wasserstein gradient only depends on the generated sample $\hat p_\theta$, and not the data distribution $\hat p$. This is a problem when the sample $\hat p_\theta$ generated by the model strongly differs from the examples coming from $\hat p$, because there is no weighting $(\alpha(\widetilde \x_n))_n$ of the generated sample that can represent the desired direction in $\Theta$. In that case, the Wasserstein gradient will point to a bad local minimum. Closeness between the two empirical samples from this optimization perspective can be ensured by adding a {\em regularization} term to the objective of the form
$$
\Omega(\theta) = \mathrm{KL}(\hat p\,\|\,p_\theta) + \eta \cdot (\|\a\|^2 + \sum_j \| \w_j \|^2).
$$
It incorporates the usual quadratic containment term, but more importantly, the KL term, that forces proximity to $\hat p$ due to the direct dependence of its gradient on it. The optimization problem becomes:
\begin{align*}
\min_\theta \quad \mathcal{W}_\gamma(p_\theta, \hat p) + \lambda \cdot \Omega(\theta)
\end{align*}
starting at point $\theta_0 = \mathrm{arg}\min_{\theta \in \Theta} \Omega(\theta)$, and where $\lambda,\eta$ are two regularization hyperparameters that must be selected. Determining the starting point $\theta_0$ is analogous to performing an initial pretraining step. Thus, the proposed Wasserstein procedure can also be seen as finetuning a standard RBM, and forcing the finetuning not to deviate too much from the initial solution.

\section{Experiments} \label{sec-experiments}

We perform several experiments that demonstrate that Wasserstein-trained RBMs learn distributions that are better from a metric perspective. First, we explore what are the main characteristics of a learned distribution that optimizes the Wasserstein objective. Then, we investigate the usefulness of these learned models on practical problems such as data completion and denoising, where the metric between observations occurs in the performance evaluation. We use two datasets: The first one is the MNIST dataset \cite{IEEE:726791:lecun98}, consisting of $60000$ handwritten digits of size $28 \times 28$. For the purpose of our experiments, the images are downsized to $14 \times 14$ pixels, and binarized with the mean pixel value of each individual pixel as threshold. We focus on modeling digits of class ``0''. There are $5923$ such examples. The second dataset is the UCI PLANTS dataset \cite{citeulike:10596116/ThePLANTSDatabase}, that associates to each plant species, a $70$-dimensional binary vector indicating its presence or absence in each US state or Canadian province. For the purpose of modeling a smooth-looking data distribution, too frequent or infrequent plants are discarded with probability $1 - e^{-(3 \cdot (\nu-0.5))^6}$ where $\nu \in [0,1]$ is the plant occurrence frequency. This results in a dataset of $6539$ plants. The two datasets are then randomly partitioned in three equal-sized subsets used for training, validation and test.

\subsection{Training, Validation and Evaluation}

All RBM models that we investigate are trained using for $\hat p_\theta$ the PCD approximation \cite{DBLP:conf/icml/Tieleman08} of $p_\theta$, where the sample is refreshed at each gradient update by one step of alternate Gibbs sampling, starting from the sample at the previous time step. We choose a PCD sample of same size as the training set ($N = \widetilde N$). The coefficients $\alpha_1,\dots,\alpha_{\widetilde N}$ occurring in the Wasserstein gradient are obtained by solving the smoothed Wasserstein dual between $\hat p_\theta$ and $p_\theta$, with smoothing parameter $\gamma = 0.1$ and distance $D(\x,\x') = \mathcal{H}(\x,\x') / \langle \mathcal{H}(\x,\x') \rangle_{\hat p}$, where $\mathcal{H}$ denotes the Hamming distance between two binary vectors. We use the centered parameterization of the RBM for gradient descent \cite{DBLP:series/lncs/MontavonM12b, DBLP:journals/neco/ChoRI13}. The learning rate is set heuristically to $0.01 (\lambda^{-1})$ during the pretraining phase and modified to $0.01 \min(1 , \lambda^{-1})$ when training on the final objective. We perform holdout validation on the quadratic containment coefficient $\eta \in \{10^{-4},10^{-3},10^{-2}\}$, and on the KL weighting coefficient $\lambda \in \{0,10^{-1},10^{0},10^{1},\infty\}$. The number of hidden units of the RBM is set heuristically to $400$ for both datasets. In our experiments, the likelihood term of the KL divergence is evaluated by estimating the partition function $Z$ using AIS with $100$ examples annealed in $1000$ steps of increasingly small temperature differences. The Wasserstein distance $\mathcal{W}_\gamma(\hat p_\theta,\hat p)$ is computed between the whole test distribution and the PCD sample at the end of the training procedure. This sample is a fast approximation of the true unbiased sample, that would otherwise have to be generated by annealing or enumeration of the states.

\begin{figure}[!t]
{\centering \sf \small
\parbox{0.49\textwidth}{\small \centering MNIST}\hfill
\parbox{0.49\textwidth}{\small \centering PLANTS}\\[-1mm]
\parbox{0.245\textwidth}{\small \centering $\text{KL}(\lambda,\eta)$}\hfill
\parbox{0.245\textwidth}{\small \centering $\mathcal{W}_\gamma(\lambda,\eta)$}\hfill
\parbox{0.245\textwidth}{\small \centering $\text{KL}(\lambda,\eta)$}\hfill
\parbox{0.245\textwidth}{\small \centering $\mathcal{W}_\gamma(\lambda,\eta)$}\\[1mm]
\includegraphics[width=0.245\textwidth]{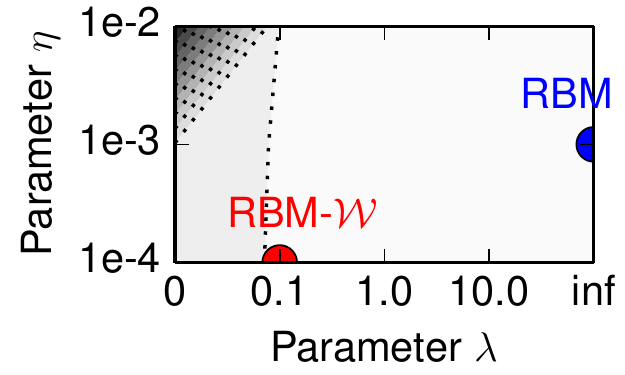}\hfill
\includegraphics[width=0.245\textwidth]{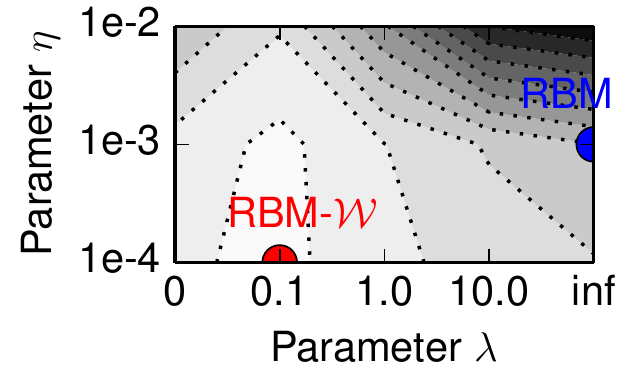}\hfill
\includegraphics[width=0.245\textwidth]{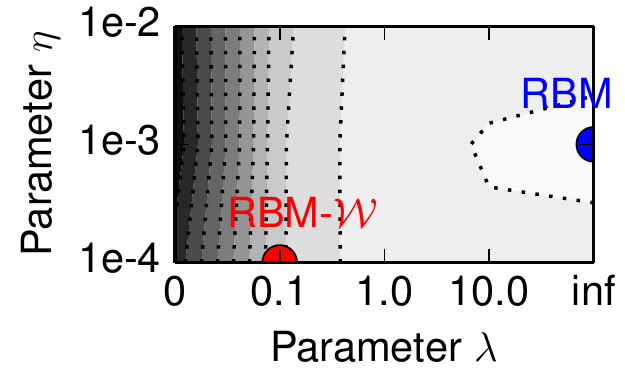}\hfill
\includegraphics[width=0.245\textwidth]{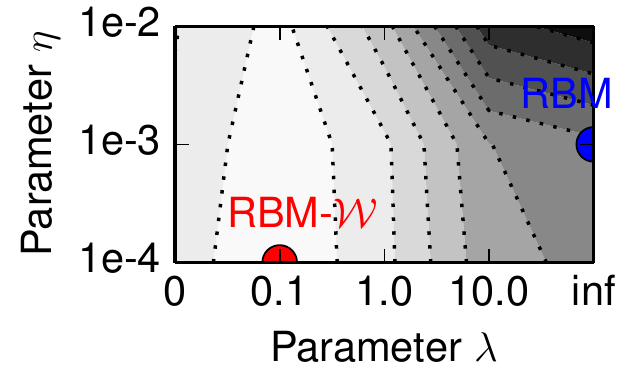}\\[-2mm]}
\caption{Contour plots showing the measure of error as a function of the regularization hyperparameters $\lambda$ and $\eta$. The best Wasserstein-trained RBMs (RBM-$\mathcal{W}$) are shown in red. The best standard RBMs (i.e.\ with $\lambda$ forced to +inf) are shown in blue.}
\label{figure:contour}
\vskip 3mm
{\centering \sf \small
\parbox{0.49\textwidth}{\small \centering MNIST}\hfill
\parbox{0.49\textwidth}{\small \centering PLANTS}\\[1mm]
\includegraphics[width=0.245\textwidth]{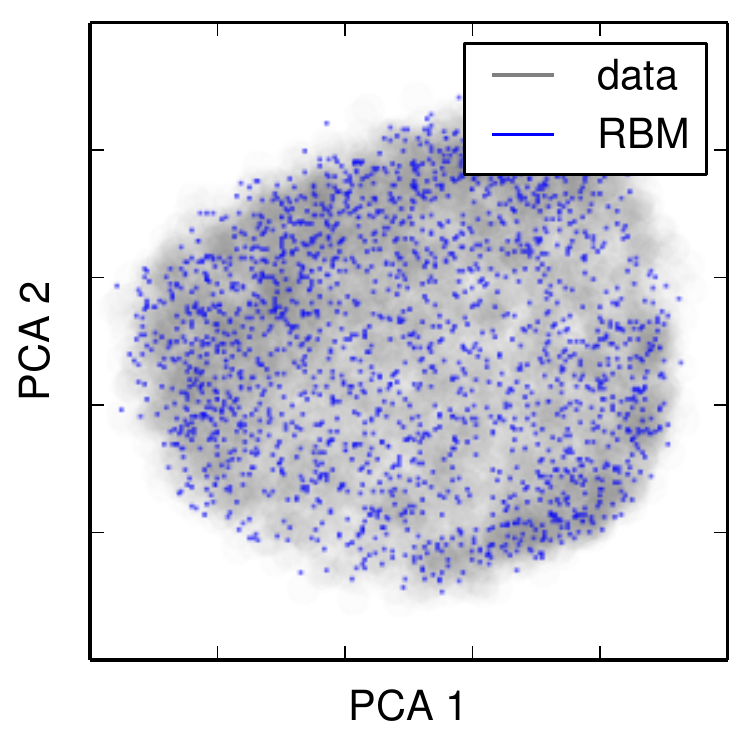}\hfill
\includegraphics[width=0.245\textwidth]{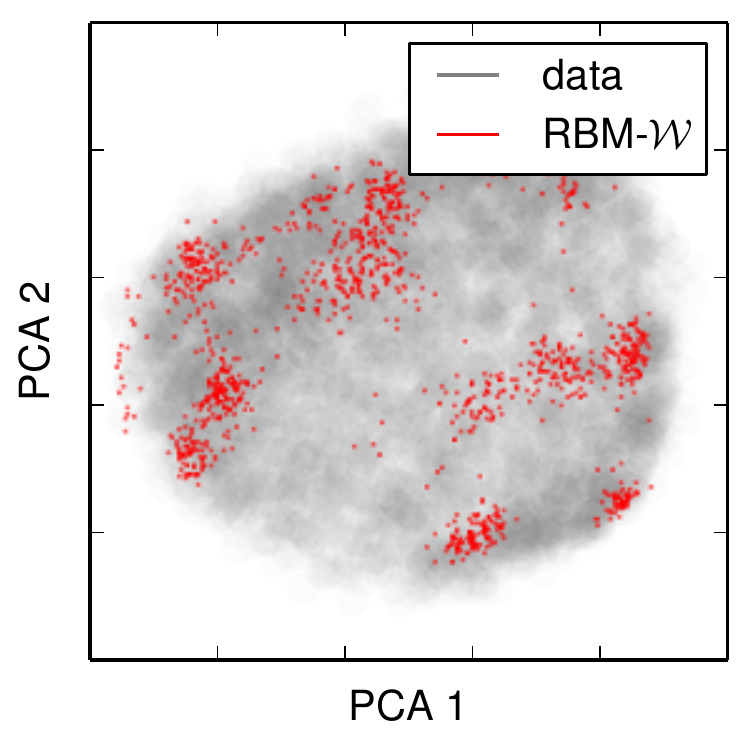}\hfill
\includegraphics[width=0.245\textwidth]{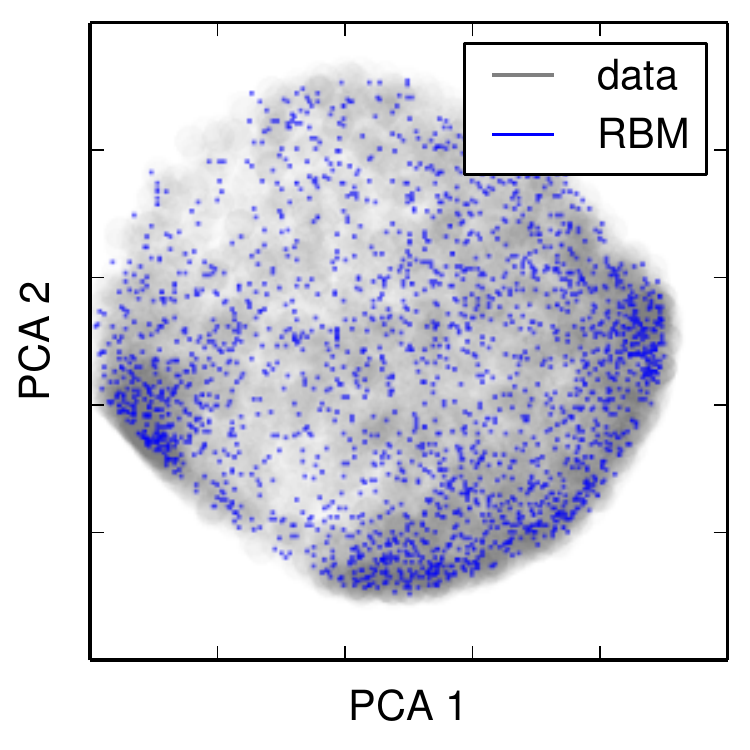}\hfill
\includegraphics[width=0.245\textwidth]{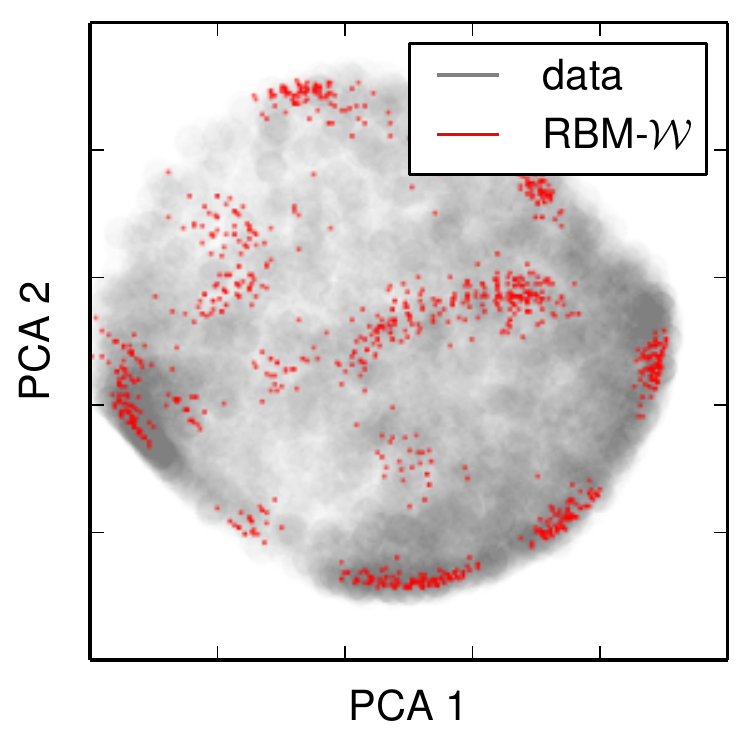}\\[-4mm]}
\caption{Two-dimensional PCA comparison of distributions learned by the RBM and the RBM-$\mathcal{W}$. Plots are obtained by projecting the learned distributions on the first components of the true distribution.}
\label{figure:pca} 
\end{figure}

\subsection{Results and Analysis}

The contour plots of Figure \ref{figure:contour} show the effect of hyperparameters $\lambda$ and $\eta$ on the KL divergence and the Wasserstein distance. For $\lambda = \infty$, only the KL regularizer is active, which is equivalent to minimizing a standard RBM. In that case, we obtain a low KL divergence. As we reduce the amount of regularization, the Wasserstein distance becomes effectively minimized and thus smaller. If $\lambda$ is chosen too small, the Wasserstein distance increases again, for the stability reasons mentioned in Section \ref{sec-training}. In all our experiments, we observed that KL pretraining was necessary in order to reach low Wasserstein distance. Not doing so leads to degenerate solutions. The relation between hyperparameters and minimization criteria is consistent across the two datasets: In both cases, the Wasserstein RBM produces lower Wasserstein distance than a standard RBM.

The PCA plots of Figure \ref{figure:pca} superimpose to the true data distribution (in gray) the distributions generated by the standard RBM (in blue) and the Wasserstein RBM (in red). In particular, the plots show the projected distributions onto the two PCA components of the true distribution. While the standard RBM distribution uniformly covers the data, the one generated by the RBM-$\mathcal{W}$ consists of a finite set of small dense clusters that are scattered across the input distribution. In other words, the Wasserstein model is biased towards these clusters, and systematically ignores other regions. Although the KL-generated distributions shown in blue may look better (the red distribution strongly departs visually from the data distribution), the red distribution is actually superior if considering the smooth Wasserstein distance as a performance metric, as shown in Figure \ref{figure:contour}.

\begin{figure}[h]
{\centering \sf \small
\begin{tabular}{c|c|c}
MNIST Data & MNIST RBM & MNIST RBM-$\mathcal{W}$\\
\includegraphics[width=0.31\textwidth]{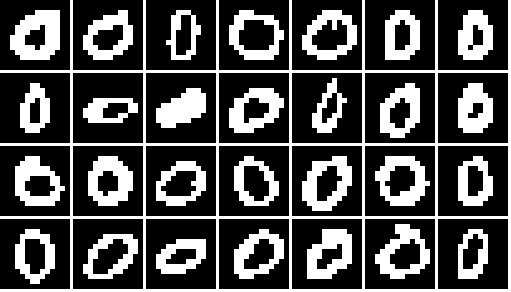} &
\includegraphics[width=0.31\textwidth]{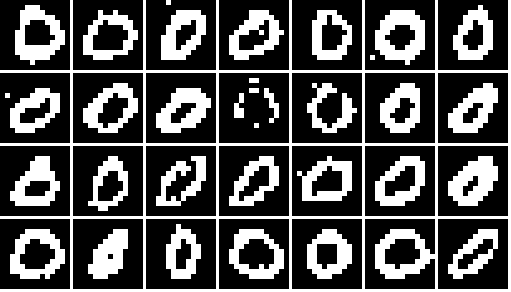} &
\includegraphics[width=0.31\textwidth]{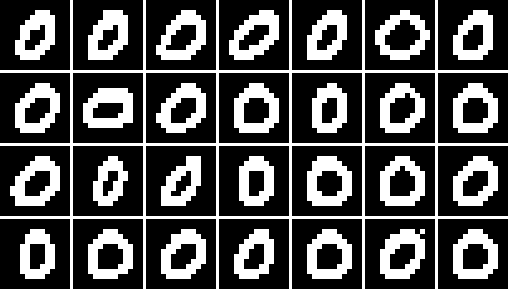}\\
\end{tabular}
\vskip 3mm
\begin{tabular}{c|c|c}
PLANTS Data & PLANTS RBM &  PLANTS RBM-$\mathcal{W}$\\
\includegraphics[width=0.31\textwidth]{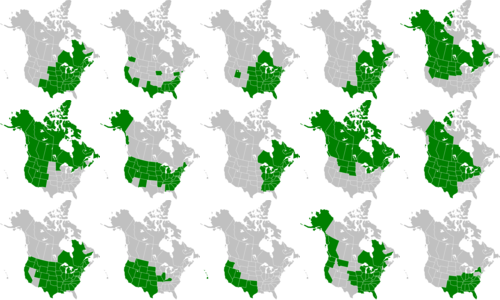} &
\includegraphics[width=0.31\textwidth]{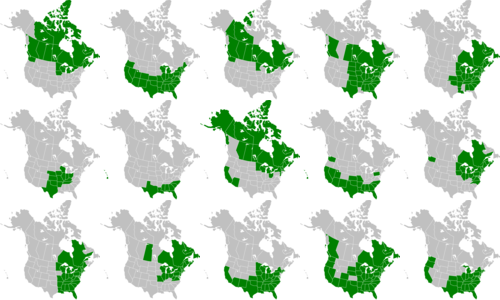} &
\includegraphics[width=0.31\textwidth]{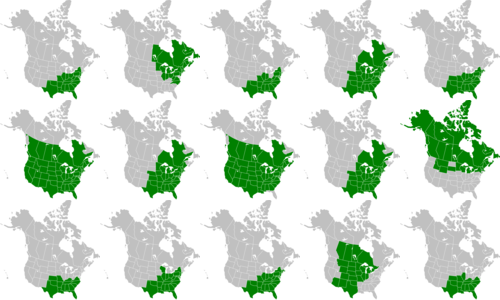}
\end{tabular}}
\caption{Samples of the MNIST and PLANTS dataset, and samples generated by the standard and the Wasserstein RBMs. (Images for the PLANTS data are automatically generated from the Wikimedia Commons template {\small \url{https://commons.wikimedia.org/wiki/File:BlankMap-USA-states-Canada-provinces.svg}} created by user Lokal\_Profil.)}
\label{figure:sample}
\end{figure}

Samples generated by the standard RBM and the Wasserstein RBM (more precisely their PCD approximation) are shown in Figure \ref{figure:sample}. The RBM-$\mathcal{W}$ produces a reduced set of clean prototypical examples, with less noise than those produced by a regular RBM. All handwritten digits generated by RBM-$\mathcal{W}$ have well-defined contours and a round shape. However, they do not reproduce the variety of shapes present in the data. Similarly, the plants species territorial spreads as generated by the RBM-$\mathcal{W}$, form compact and contiguous regions that are prototypical of real spreads, but are also less diverse than the data or the sample generated by the standard RBM.

\subsection{Application to Data Completion and Denoising}

In order to demonstrate the practical relevance of Wasserstein distance minimization, we apply the learned models to the task of data completion and data denoising, for which the use of a metric is crucial: Data completion and data denoising performance is generally measured in terms of \emph{distance} between the true data and the completed or denoised data (e.g. Euclidean distance for real-valued data, or Hamming distance $\mathcal{H}$ for binary data). Remotely located probability mass that may result from simple KL minimization would incur a severe penalty on the completion and denoising performance metric. Both tasks have useful practical applications: Data completion can be used as a first step when applying discriminative learning (e.g. neural networks or SVM) to data with missing features. Data denoising can be used as a dimensionality reduction step before training a supervised model. Let the input $\x = [\v,\h]$ be composed of $d-k$ visible variables $\v$ and $k$ hidden variables $\h$.

\begin{figure}[h]
\centering
\includegraphics[width=1.0\textwidth]{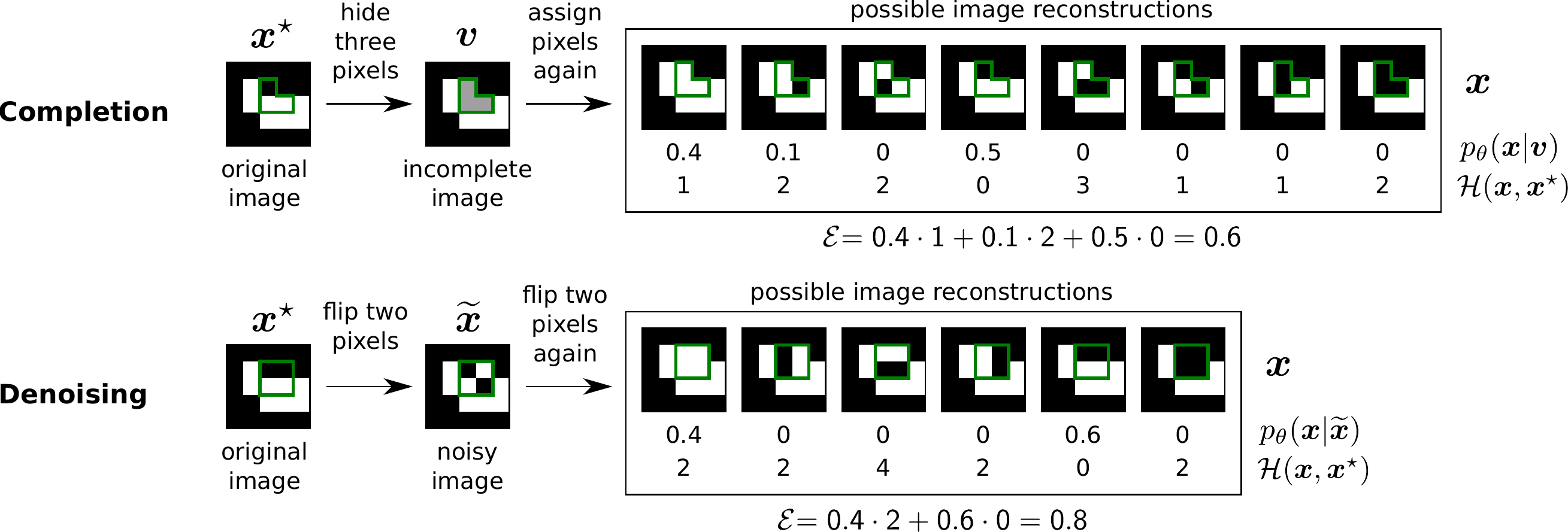}\vskip -2mm
\caption{Illustration of the completion and denoising setup. For each image, we select a known subset of pixels, that we hide (or corrupt with noise). Each possible reconstruction has a particular Hamming distance to the original example. The expected Hamming error is computed by weighting the Hamming distances by the probability that the model assigns to the reconstructions.}
\label{figure:denoising}
\vskip 3mm
{\centering \sf \small
\begin{tabular}{cc}
\parbox{0.235\textwidth}{\centering \small Completion (MNIST)}
\parbox{0.235\textwidth}{\centering \small Completion (PLANTS)}&
\parbox{0.235\textwidth}{\centering \small Denoising (MNIST)}
\parbox{0.235\textwidth}{\centering \small Denoising (PLANTS)}\\[1mm]
\includegraphics[width=0.235\textwidth]{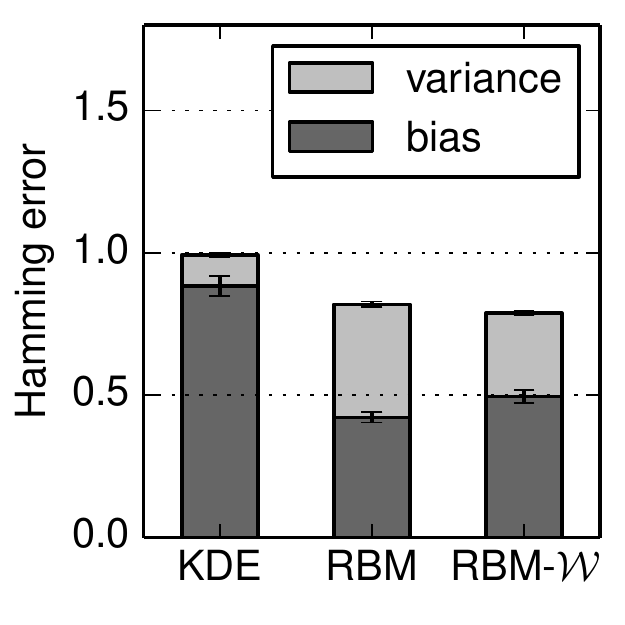}
\includegraphics[width=0.235\textwidth]{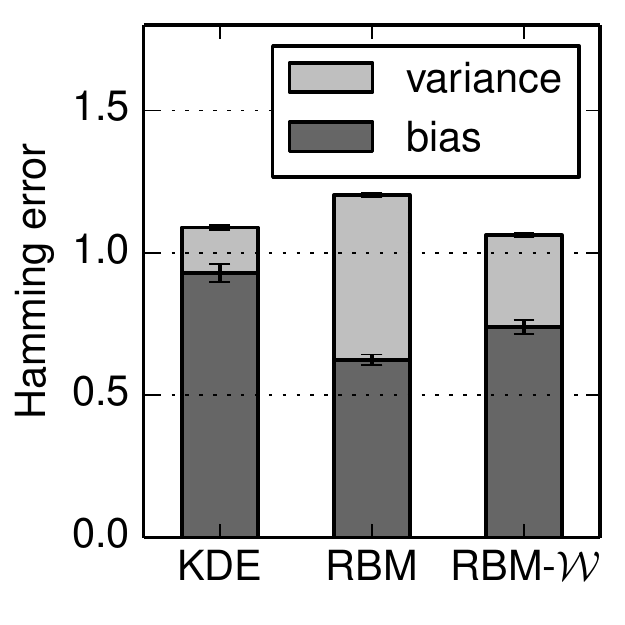}&
\includegraphics[width=0.235\textwidth]{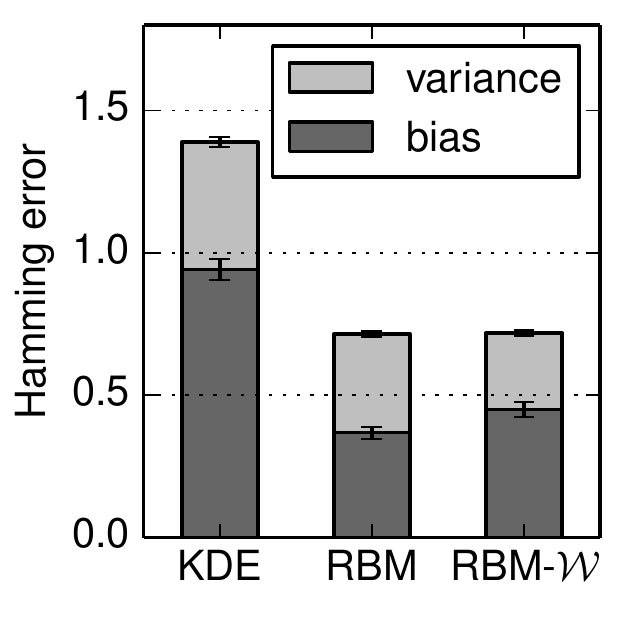}
\includegraphics[width=0.235\textwidth]{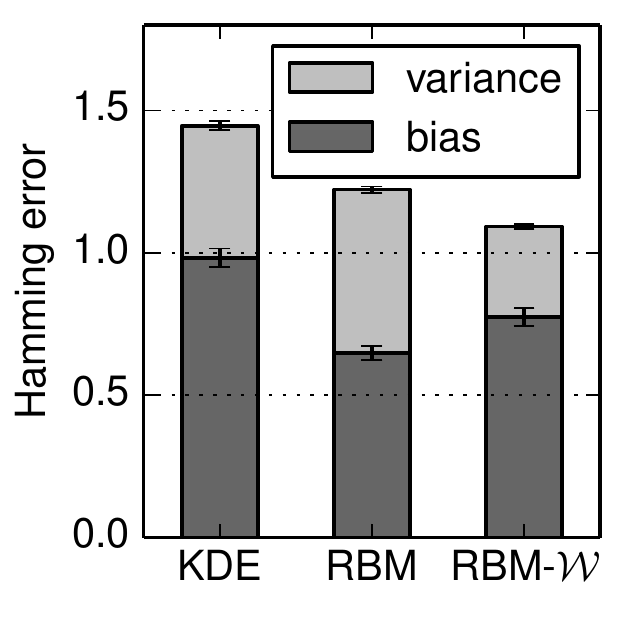}\\[-5mm]
\end{tabular}}
\caption{Performance on the completion and denoising tasks of the kernel density estimation, the standard RBM and the Wasserstein RBM. The total length of the bars is the expected Hamming error. Dark gray and light gray sections of the bars give the bias-variance decomposition.}
\label{figure:bars}
\end{figure}

\paragraph{Data Completion} The setting of the data completion experiment is illustrated in Figure \ref{figure:denoising} (top). The distribution $p_\theta(\x|\v)$ over possible reconstructions can be sampled from using an alternate Gibbs sampler, or by enumeration. The expected Hamming distance between the true state $\x^\star$ and the reconstructed state modeled by the distribution $p_\theta(\x|\v)$ is given by iterating on the $2^k$ possible reconstructions:
$
\mathcal{E} = {\textstyle \sum_{\h \in \{0,1\}^k}} ~p_\theta(\x \,|\, \v) \cdot \mathcal{H}(\x,\x^\star).
$
Since the reconstruction is a probability distribution, we can compute the expected Hamming error, but also its bias-variance decomposition. On MNIST, we hide randomly located image patches of size $3 \times 3$ (i.e. $k=9$). On PLANTS, we hide random subsets of $k = 9$ variables. Results are shown in Figure \ref{figure:bars} (left), where we compare three types of models: Kernel density estimation (KDE), standard RBM (RBM) and Wasserstein RBM (RBM-$\mathcal{W}$). The KDE estimation model uses a Gaussian kernel, with the Gaussian scale parameter chosen such that the KL divergence of the model from the validation data is minimized. The RBM-$\mathcal{W}$ is better or comparable the other models. Of particular interest is the structure of the expected Hamming error: For the standard RBM, a large part of the error comes from the variance (or entropy), while for the Wasserstein RBM, the bias term is the most contributing. This can be related to what is observed in Figure \ref{figure:pca}: For a data point outside the area covered by the red points, the reconstruction is systematically redirected towards the nearest red cluster, thus, incurring a systematic bias.

\paragraph{Data Denoising} Here, we consider a simple noise process where for a predefined subset of $k$ variables, denoted by $\h$ a known number $l$ of bits flips occur randomly. Remaining $d-k$ variables are denoted by $\v$. The setting of the experiment is illustrated in Figure \ref{figure:denoising} (bottom). Denoting $\x^\star$ the original and $\widetilde \x$ its noisy version resulting from flipping $l$ variables of $\h$, the expected Hamming error is given by iterating over the $\binom{k}{l}$ states $\x$ with same visible variables $\v$ and that are at distance $l$ of $\widetilde \x$:
$
\mathcal{E} = {\textstyle \sum_{\h \in \{0,1\}^k}} ~p_\theta(\x \,|\, \v, \mathcal{H}(\x,\widetilde \x) = l) \cdot \mathcal{H}(\x,\x^\star) .
$
Note that the original example $\x^\star$ is necessarily part of this set of states under the noise model assumption. For the MNIST data, we choose randomly located images patches of size $4 \times 3$ or $3 \times 4$ (i.e. $k=12$), and generate $l = 4$ random bit flips within the selected patch. For the PLANTS data, we generate $l = 4$ bit flips in $k = 12$ randomly preselected input variables. Figure \ref{figure:bars} (right) shows the denoising error in terms of expected Hamming distance on the same two datasets. The RBM-$\mathcal{W}$ is better or comparable to other models. Like for the completion task, the main difference between the two RBMs is the bias/variance ratio, where again the Wasserstein RBM tends to have larger bias. This experiment has considered a very simple noise model consisting of a fixed number of $l$ random bit flips over a small predefined subset of variables. Denoising highly corrupted complex data will however require to combine Wasserstein models with more flexible noise models such as the ones proposed by \cite{DBLP:conf/cvpr/TangSH12}.

\section{Conclusion} \label{sec-conclusion}

We have introduced a new objective for Boltzmann machines based on the smooth Wasserstein distance. Unlike the usual Kullback-Leibler (KL) divergence, our objective takes into account the metric of the data. The objective admits a simple gradient, that can be computed by solving the dual of the Wasserstein distance between the learned and observed distributions. We learned a Wasserstein model on two simple problems: In both cases, the learned distributions strongly departed from the KL model, and formed instead a set of clusters of prototypical examples (well-shaped digits for MNIST, and contiguous territorial spreads for PLANTS).

We have evaluated the Wasserstein RBM on two basic completion and denoising tasks, for which the metric of the data intervenes in the performance evaluation. In this simple setting, we have demonstrated the superiority of the RBM-$\mathcal{W}$ over the standard RBM, and how the bias-variance structure of the estimator systematically differs. Our contribution aims principally at introducing a novel type of objective for the Boltzmann machine where it did not exist before, and showing that Boltzmann machines and Wasserstein methods can be combined. In particular, our work gives an additional practical motivation for developing Wasserstein methods that run quickly on large datasets.

\paragraph{Acknowledgments} This work was supported by the Brain Korea 21 Plus Program through the National Research Foundation of Korea funded by the Ministry of Education. This work was also supported by the grant DFG (MU 987/17-1). M. Cuturi gratefully acknowledges the support of JSPS young researcher A grant 26700002. Correspondence to GM, KRM and MC.

\bibliographystyle{plain}
\bibliography{wassersteinboltzmann}

\end{document}